\documentclass{article}

\usepackage[utf8]{inputenc}
\usepackage{amsmath}
\usepackage{amssymb}    % Alex
\usepackage{amsthm}     % Alex
\usepackage[affil-it]{authblk} % affiliations for authors on title page
\usepackage{array} % Needed for using > {\raggedright\arraybackslash}
\usepackage{booktabs}
\usepackage{caption}
\usepackage{csquotes}   % Alex
\usepackage{float}
\usepackage{hyperref}
\usepackage{cleveref}
\usepackage{marvosym}
\usepackage{mathtools}  % Alex
\usepackage{nicefrac}
\usepackage{tabularx}
\usepackage{tikz}       % Alex
\usepackage{siunitx}
\usepackage{xspace}
\usepackage{xifthen}
\usepackage{wasysym}
\usepackage{subcaption}
\usepackage{xcolor}

\usepackage[a4paper,
            bindingoffset=0.2in,
            left=1in,
            right=1.375in,
            top=1in,
            bottom=1in,
            footskip=.25in]{geometry}

\usepackage[textsize=tiny]{todonotes}
\usepackage[backend=biber,style=numeric,sorting=nty,isbn=false,url=false,eprint=false,citestyle=authoryear-comp]{biblatex}
\usepackage{pgfplots}

\graphicspath{{images/}}

\newcommand{\ite}{i.\,e.\xspace}
\newcommand{\eg}{e.\,g.\xspace}
\newcommand{\cf}{cf.\xspace}
\newcommand{\etc}{etc.\xspace}

\newif\ifcolorversion
\colorversiontrue % Turn this line into a comment to switch to white and black version

\newcommand{\ifcolor}[2]{
    \ifcolorversion
        #1
    \else
        #2
    \fi
}

\bibliography{references}

\begin{document}

\title{Does the Brain Infer Invariance Transformations from Graph Symmetries?}

\author{Helmut Linde}
\affil{Merck~KGaA, Darmstadt, Germany}
\affil{Transylvanian~Institute~of~Neuroscience, Cluj-Napoca, Romania}

\maketitle

\begin{abstract}
The invariance of natural objects under perceptual changes is possibly encoded in the brain by symmetries in the graph of synaptic connections. The graph can be established via unsupervised learning in a biologically plausible process across different perceptual modalities. This hypothetical encoding scheme is supported by the correlation structure of naturalistic audio and image data and it predicts a neural connectivity architecture which is consistent with many empirical observations about primary sensory cortex.
\end{abstract}

\section{Introduction}\label{sec:Introduction} 

To build an internal model of the environment, the brain needs some representation of \emph{invariance transformations} which may change the way how one and the same object is perceived.
Such transformations include, for example, translations, rotations, or rescaling of visual objects, as well as a change of key or octave in music.
They are employed both passively, as in the quick recognition of rotated letters \parencite{corballis_decisions_1978}, and actively, as in mental rotation tasks \parencite{shepard_mental_1971}. 
Mastering such invariance transformations is instrumental in the abstraction process of untangling perceptual input into different mental categories like object class, its location and orientation in space, or its size.

The central claim of this article is that the brain learns and encodes invariance transformations in graph symmetries as a substrate for computational processes.
We construct a theory on the assumption that invariances initially manifest themselves as approximate symmetries of the probability distribution on the space of all possible perceptual observations. This distribution gives rise to a set of feature detectors via some unsupervised learning process, as it may be the case in primary sensory cortices. We assume that the set of detectors \enquote{inherits} a symmetry transformation of the distribution in the sense that every feature detector implies the existence of another detector for the transformed feature. In that case the invariance transformation can be expressed as a permutation of the feature detectors. Importantly, their pairwise activity correlations over many observations are then invariant under said permutation. Assuming a process of Hebbian learning to shape the recurrent synaptic connections between the feature detectors, those correlations are reflected in the synaptic weights. The invariance transformation should therefore give rise to a symmetry in the graph of synaptic connections between the feature detectors.

This article is structured as follows: In Section \ref{sec:model}, we formalize the ideas outlined above by defining the problem of learning invariance transformations with mathematical precision and demonstrating how a neural connectivity pattern can naturally develop to encode these invariances. Section \ref{sec:empirical} summarizes empirical evidence supporting two key points: (1) the statistics of natural stimuli align with the assumptions made in this study, and (2) neural connectivity patterns in the neocortex are consistent with the predictions of our theory.

Section \ref{sec:discussion} explores potential extensions of the proposed theory, addresses possible objections, and compares it to alternative models for learning perceptual invariances. Additionally, this section outlines experiments designed to evaluate the theory against real-world observations. Finally, Section \ref{sec:conclusion} offers a summary and conclusion of the article.

%-------------------------------------------------------
\section{A New Way of Transformation Learning} \label{sec:model} %
%-------------------------------------------------------
%-------------------------------------------------------
\subsection{Problem Statement} \label{sec:model:problem} % 
%-------------------------------------------------------
% Assumption: some stimuli generating process, can be modeled as drawing froma random distribution
% invariance transformation is a symmetry of this random distribution
% so the task is to find symmetries of a random distribution given some samples (observations)
% problem: distribution is in a very high dimensional space
% idea: projection of this distribution to lower dimensional spaces!

Consider a model brain which perceives its environment over some extended time period via a set of $n$ information channels.
Each of them might represent one atomic percept (like a pixel of an image) or some pre-aggregated combination thereof (like a small edge in an image).
We call each such information channel a \emph{feature detector}, independent of whether it represents an atomic percept or pre-aggregated ones.

An \emph{observation} is a snapshot of the feature detectors' state at some time $t$. For simplicity, we assume that at any time a feature can only be either present or not, represented by the numbers $1$ and $0$.

We model the process of perception as a repeated independent random drawing from a discrete probability distribution $\Psi: \{0;1\}^n \rightarrow [0;1]$ on the feature state space, where $n$ is the total number of feature detectors under consideration.
Let now $T$ be a transformation on the feature state space which acts via a permutation of the features.
We call $T$ an \emph{invariance transformation} if it has no effect on $\Psi$, \ite if $\Psi(Tx) = \Psi(x)$ for all $x \in \{0;1\}^n$. 

How can one find the invariance transformations of $\Psi$ given a finite set of observations $x$ only?
In practically relevant cases, this problem is hard because $\Psi$ is a probability distribution in high-dimensional space and to approximate $\Psi$ its value needs to be estimated on a number of points which grows exponentially with the number of dimensions.
In the following we shall discuss an approach to identify candidates for invariance transformations without being able to reconstruct $\Psi$ explicitly. Since it relies heavily on \emph{graph symmetries}, a brief general introduction to that concept is in order before introducing the approach.

%-------------------------------------------------------
\subsection{Graph Symmetries} \label{sec:model:symmetry} %
%--------------------------------------------------------

\begin{figure}[tb]
    \centering
    \includegraphics[width=2.5in]{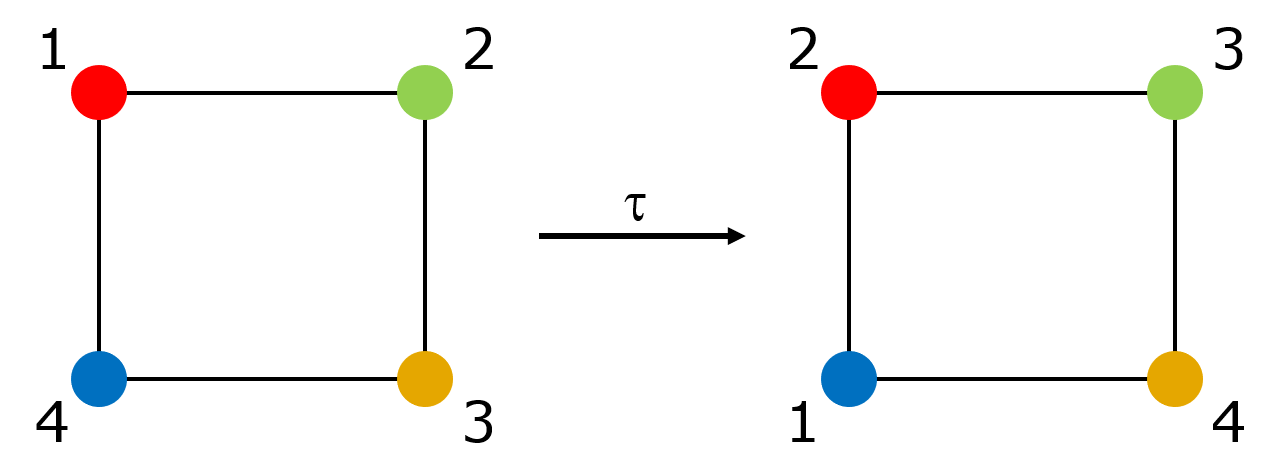}
    \caption{%
        Simple example of graph symmetries. Assume that the four edges have equal weights. Then, the cyclic permutation $\tau$, relabeling each node by the next higher number and 4 by 1, is a graph automorphism. Together with $\tau^2$, $\tau^3$, and the identity $\tau^4$, it characterizes a symmetry (\enquote{rotation}) of the graph. Similarly, the exchange of nodes 1 and 3 is an automorphism which gives rise to another symmetry (\enquote{reversal}). Further automorphisms can be constructed by combining the two symmetries. An exchange of the nodes 1 and 2, on the other hand, is not an automorphism since it breaks the link between $1$ and $4$, inter alia.
    }
    \label{fig:symmetry}
\end{figure}

A \emph{graph} is a collection of uniquely identifiable (\enquote{labeled})  nodes, each pair of which can be connected by an edge, and it is called \emph{weighted} if each edge is associated with a number.
Permutations can act on a graph by interchanging node labels without altering the edges.
If the relabeled graph is structurally identical to the original one, the permutation is called a \emph{graph automorphism}.
More formally, the permutation $\tau$ is an automorphism if and only if for every pair of nodes $u$ and $v$ either the edge between $u$ and $v$ has the same weight as the one between $\tau(u)$ and $\tau(v)$, or neither of the two edges exists.
Further automorphisms can be constructed by applying $\tau$ repeatedly.
This amounts to a chain of node exchanges which will ultimately, after a certain number of steps, restore the original graph.
Such a sequence of permutations characterizes one \emph{graph symmetry} of $G$.  \Cref{fig:symmetry} illustrates these definitions with a simple example.

%-------------------------------------------------------
\subsection{The Concurrence Graph} \label{sec:model:projections} %
%--------------------------------------------------------

The crucial idea to find invariance transformations is to replace $\Psi$ by some of its projections to lower-dimensional spaces: For example, the marginal distribution of an individual component $x_k$ or of two components $x_k$ and $x_{k'}$ can be estimated accurately from a relatively small number of observations. The marginal distributions \enquote{inherit} the symmetries of the original distribution, \ite if $\Psi$ is unaltered by some permutation $\tau$ of the feature detectors, then so are the marginal distributions (see the appendix for details). Yet the reverse statement is not true: Certain marginal distributions of $\Psi$ might exhibit a symmetry property which $\Psi$ itself does not have. For example, if visual perceptions are translation invariant, and if we choose the features to be individual pixels, then every $\Psi_k$ (\ite the probability of a pixel $k$ being \enquote{on} or \enquote{off}) is equal. Thus every possible permutation of pixels leaves these marginal distributions of $\Psi$ unchanged, which is obviously not true for $\Psi$ itself. In the light of these considerations, it is necessary to find a reasonable trade-off when choosing the projections applied to $\Psi$: On the one hand, the dimension of the projected functions should be small enough that they can be approximated by the observations given. On the other hand it should not be so small that too many additional symmetries appear as artifacts.

For the present discussion, the projection of $\Psi$ to two-dimensional spaces is the most relevant case. For every pair $k \neq m$ of features, the projected probability distribution $\Psi_{k,m}$ is determined by three numbers: The individual probabilities of the features $k$ and $m$ to be \enquote{on} and their joint probability to be \enquote{on} simultaneously. 

Note the special case where all the $\Psi_k$ distributions are known and equal, \ite every feature has the same probability of being \enquote{on}: the distribution $\Psi_{k,m}$ is then determined by only one number for each pair $\{k; m\}$, namely by the probability of features $k$ and $m$ being \enquote{on} simultaneously.
If we now associate each feature with the node of a graph, and take this probability to be the weight of the edge between nodes $k$ and $m$, then \textbf{every invariance transformation of $\Psi$ turns into a symmetry of this graph}.
Since the edges of the graph are a measure for the probability of two features being \enquote{on} jointly, we shall call it the \emph{concurrence graph}.

For the present purpose, we are only interested in symmetries of the concurrence graph and not in the numeric values of its edge weights. In particular, it does not matter if the weight between $k$ and $m$ represents the probability of those features being \enquote{on} simultaneously or some strictly monotonous function of that probability. This flexibility strengthens the biological plausibility of the presented concept and we shall therefore use the term \enquote{concurrence graph} in a broader sense for any graph whose edge weights are computed by some strictly monotonous function of the actual probabilities. 

In summary, the concurrence graph is defined as follows:
\begin{enumerate}
    \item Each feature detector can be identified with one node of the concurrence graph.
    \item The recurrent synaptic connections between the feature detectors correspond to the edges of the concurrence graph.
    \item The number of observations during development which activated a certain pair of feature detectors simultaneously determines the weight of the edge between them in a strictly monotonous way.
\end{enumerate}

In the following, we will discuss how the concurrence graph can form organically in the synaptic connections of a neural system.

%-------------------------------------------------------
\subsection{Formation of the Concurrence Graph in the Synaptic Structure} \label{sec:model:concurrencegraph} % 
%-------------------------------------------------------

% Feature detectors form, they tend to form in such a way that their fire probabilities are equal. correlationgraph forms
% a symmetry of the stimulus generation function translates to a graph symmetry
% the symmetries of Psi are now encoded in graph symmetries and can inpricipele be accessed by algorithm running on this neural substrate.

\begin{figure}[tb]
    \centering
    \includegraphics[width=2.5in]{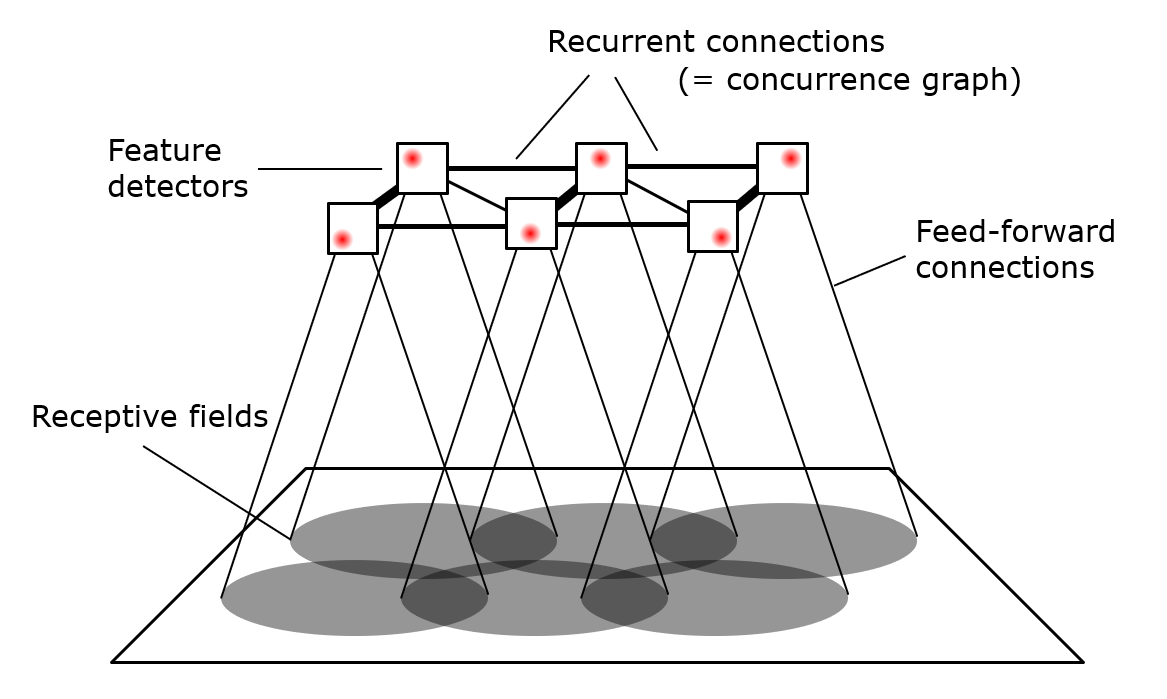}
    \caption{%
        According to the postulated theme of connectivity for primary sensory cortices, feature detector units are shaped by external stimuli transmitted through feed-forward connections in a process of competitive Hebbian learning. Recurrent connections are then built up also according to a Hebbian learning rule, thus becoming a measure for the correlation between two feature detectors. 
    }
    \label{fig:theme}
\end{figure}

According to the present proposal, the concurrence graph is part of a \emph{theme of connectivity} (see \Cref{fig:theme}) which forms organically in primary sensory cortex via biologically plausible mechanisms.

In the subsequent description of this mechanism, we purposefully refrain from specifying whether the graph materializes at the level of singular neurons or collective entities. In other words, a \enquote{feature detector} constitutes a computational unit that can be conceived as a single neural cell or, for example, an entire (micro)column. Correspondingly, the edge weights within the concurrence graph represent either the strength of a single synapse or a measure of the total connectivity between two aggregates of cells.

The first phase of the process is the formation of the feature detectors through competitive Hebbian learning \parencite{rumelhart_feature_1985}. Each computational unit receives direct feed-forward perceptual input and whenever it is activated strongly enough, it will \enquote{fire}, inhibit the other units, and tune its feed-forward synaptic connections closer to the current input in accordance with the Hebbian learning rule. Examples for such competitive learning algorithms are Kohonen Maps \parencite{kohonen_self-organized_1982}, (Growing) Neural Gas \parencite{martinetz_topology_1994} or variants of sparse coding dictionary learning \parencite{Elad:2010}. While the quantitative details may differ between these implementations of competitive learning, the qualitative outcome tends to be similar: Ultimately each unit represents a certain pattern in sensory perception.

In the second phase of the learning process, which may overlap in time with the first phase, the recurrent connections between the feature detectors are established. Simply applying Hebbian learning again, the synapses between two feature detectors are strengthened whenever they are both activated simultaneously.

Consequently, the concurrence graph has materialized in the structure of recurrent neural connections between the feature detectors.
Its symmetries represent the invariances of the environmental stimuli that have shaped the network during the training process.
It can now serve as a substrate on which different neural algorithms can be implemented, like object classification, mental imagery tasks, or the planning of bodily motions \parencite{powell_can_2021}.
In particular, when a new object is perceived for the first time, it is decomposed into the same set of features that form the graph's nodes, and its symmetries can directly be applied to the new object.

\subsection{Visualizing the concurrence graph} \label{sec:model:visualizing}

\begin{figure}[tb]
    \centering
    \includegraphics[width=2.5in]{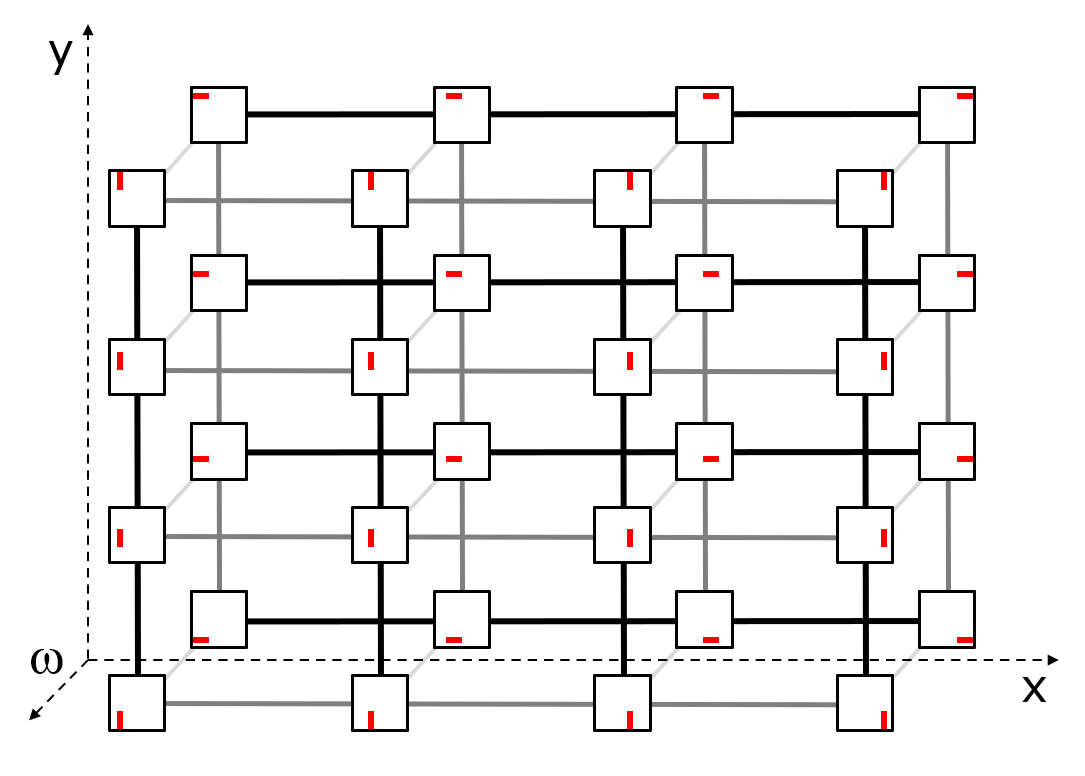}
    \caption{%
        Schematic of a concurrence graph. Nodes are boxes, each of which corresponds to a feature detector with orientation $\omega$ on an image patch at the respective $(x,y)$ position. The line strength between boxes represents the weight of the respective edge, with connections between co-linear features being particularly strong. Lines between diagonal or indirect neighbors are omitted for better readability. The graph is symmetric under reflections and under translations except for border effects. Another symmetry is a \ang{90} rotation of the $x$-$y$-plane combined with a simultaneous interchange of horizontal and vertical feature detectors.
    }
    \label{fig:graph}
\end{figure}

\begin{figure}[tb]
    \centering
    \includegraphics[width=2.5in]{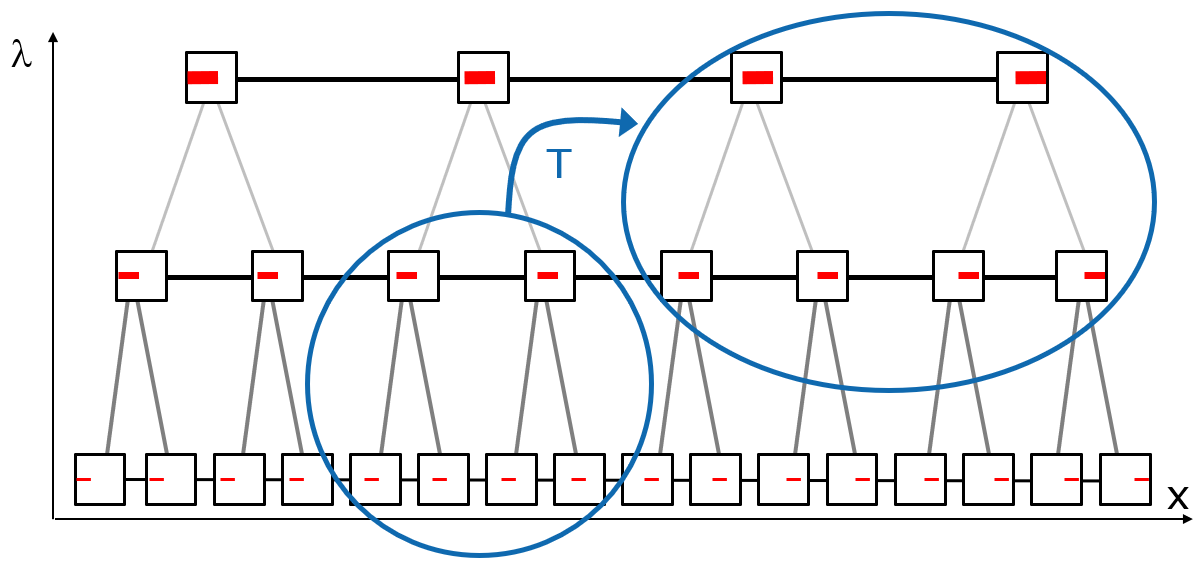}
    \caption{%
        Concurrence graph as in \Cref{fig:graph}, depicted with focus on scale invariance. A subset of feature detectors with receptive fields of different sizes $\lambda$ are shown. The organic development of scale invariant detector sets is hypothetically a consequence of the corresponding invariance of natural visual stimuli, \cf \Cref{sec:empirical:natural}. Despite the suggestive hierarchical structure all depicted connections are recurrent, while feed-forward connections to the detectors are omitted. The transformation $T$ combines a translation in $x$ and a rescaling. It approximates a graph automorphism, characterized by its action on the nodes in\ifcolor{blue}{}ovals and a corresponding action on all other nodes, except for the \enquote{border effects} at the extreme values of $x$ and at the highest and lowest scaling levels.
    }
    \label{fig:scaling}
\end{figure}

Simple schematics of a concurrence graph with a caricature of V1 cells are shown in \Cref{fig:graph,fig:scaling}.
The former shows a subset appropriate to exhibit translation and rotation invariance, while the latter demonstrates scale invariance.
For readability, only a very small number of feature detectors are depicted in each figure.
 
Naturally, a graph that accurately models invariance learning in V1 is substantially more complex than the two figures presented, as it must incorporate a node for each conceivable feature detector. This encompasses a minimum of a four-dimensional space of feature detectors, tuned to varying retinotopic positions, orientations, and scales. In particular, a more comprehensive schematic would not be limited to features with horizontal or vertical orientations, as depicted in Figure \ref{fig:graph}, but would also incorporate numerous intermediate angles.

Visualizing such a graph in a manner that clearly conveys all four dimensions and the corresponding graph symmetries is a formidable challenge. Nevertheless, it is well-established that these feature detectors do exist in the brain and form a highly interconnected network (refer to Section \ref{sec:empirical:connectivity}). Based on the theoretical considerations presented thus far, it is reasonable to anticipate that even this extensive graph possesses the relevant symmetries.

An example of how the concurrence graph conceptually supports the completion of perceptual tasks is given in \Cref{fig:intuitive}.
The\ifcolor{blue and the green}{two}\enquote{H} are connected by a graph symmetry, as explained in the caption.
Applied to this example, the claim of the present article reads as follows: The brain considers the two letters \enquote{the same} because of their indistinguishable embedding in the graph structure.

Note that the concurrence graph in \Cref{fig:intuitive} could have been formed by visual experience, even without prior exposure to the specific letter "H". This exemplifies how the symmetries of the concurrence graph encode invariance transformations, independent of the stimuli from which these transformations were initially learned. Furthermore, it demonstrates that these transformations can be readily applied to novel stimuli.

\begin{figure}
    \hfill
    \begin{subfigure}{1.8in}
        \centering
        \includegraphics[width=1.8in]{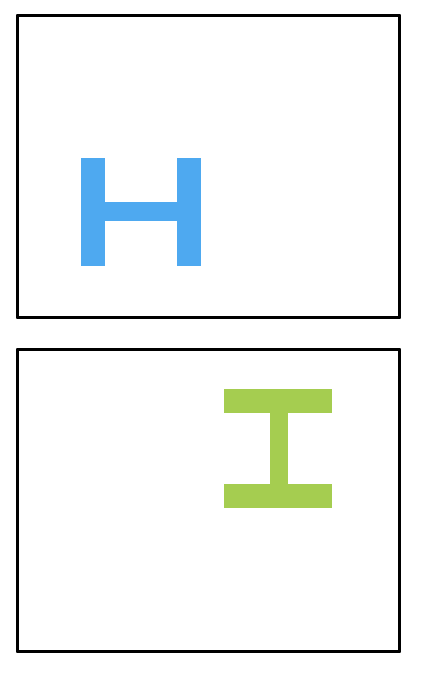}
        \caption{Visual stimuli}
        \label{fig:intuitive:stimuli}
    \end{subfigure}
    \hfill
    \begin{subfigure}{2.6in}
        \centering
        \includegraphics[width=2.6in]{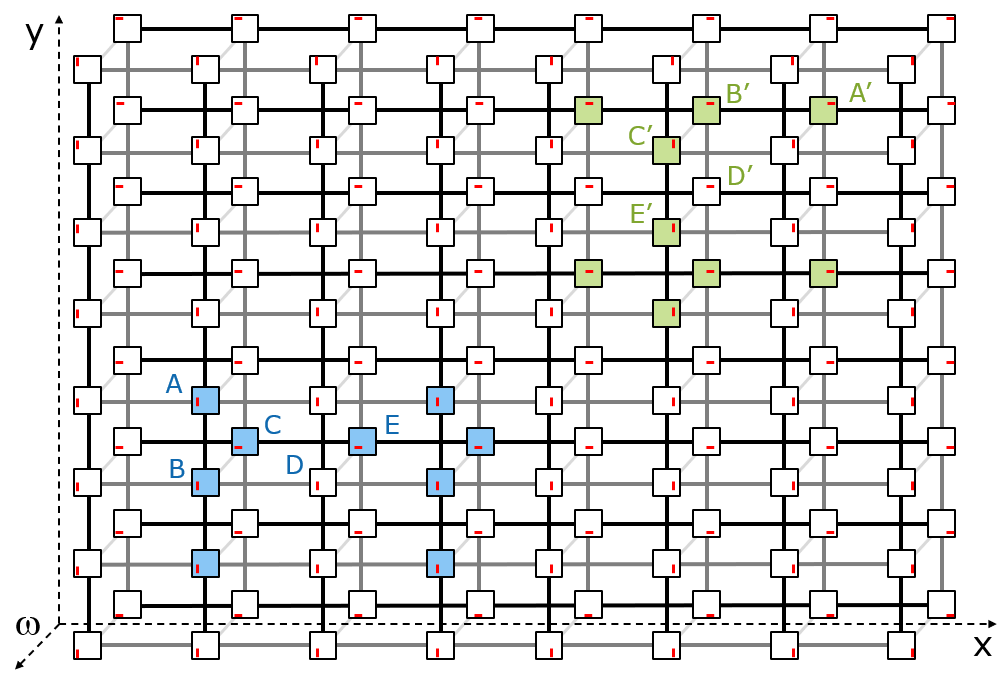}
        \caption{Concurrence graph}
        \label{fig:intuitive:graph}
    \end{subfigure}
    \caption{Two visual stimuli (\subref{fig:intuitive:stimuli}) activate sets of feature detectors (\subref{fig:intuitive:graph}) marked in\ifcolor{the corresponding color (note that this is a model of black-and-white vision and the colors are for differentiation only)}{grey}. The two stimuli are connected by an invariance transformation, namely a translation and a rotation by $\ang{90}$. The corresponding transformation in the concurrence graph is a permutation of all feature detectors, in particular mapping $A$ to $A'$, $B$ to $B'$, \etc It is also a graph symmetry: Note how the\ifcolor{blue and the green}{two sets of}nodes are identically embedded in the graph. For example, A is connected to B by a strong black line just like A' to B', and analogously for all other corresponding pairs of nodes (including those which are not activated by the stimuli, like D).}
    \label{fig:intuitive}
\end{figure}

%---------------------------------------------
\subsection{Connecting the model to reality} \label{sec:model:connecting} % 
%-------------------------------------------------------

We conclude \Cref{sec:model} with a discussion of how to interpret the proposed model in relation to biological neural networks.

As stated above, the concept of the concurrence graph and its formation is agnostic about the precise nature of the feature detectors.
Each of them could be a single neuron or a cluster of cells with similar receptive characteristics.
While synaptic connections between individual cells make learning possible, the presence of a synapse between two neurons is presumably somewhat random, especially if they are not in close proximity.
This imposes a limitation on the precision of graph symmetries on the single neuron level.
However, when considering aggregated groups of cells, the impact of stochastic effects lessens, and graph symmetries should become more pronounced.

A crucial component of biological neural networks not yet addressed is the presence of inhibition. The latter is believed to regulate overall neural activity and may have a significant computational function, although its exact nature remains unclear. In the visual cortex, for example, inhibition can be selective to the orientation of features relative to each other \parencite{Angelucci:2017}.

In the proposed model, inhibition plays a role in both phases of graph formation. In the first phase, winner-takes-all dynamics enable each computational unit to concentrate on learning a specific feature. This implicitly relies on an inhibitory mechanism that suppresses non-winning units for each presented stimulus. In the second phase, mutual inhibition may modulate the frequency at which two neurons fire together, thus affecting the weights during graph formation. Since inhibitory mechanisms are intricate and multifactorial, we do not explicitly incorporate them into the model. Even if they were considered in the model, they would not break the process of encoding the invariance transformations in graph symmetries, as long as the inhibition mechanism itself is invariant under these symmetries. The latter is a plausible assumption since the inhibition is mediated via neural connections which, as we have seen, exhibit these very symmetries.

Additionally, the model could be expanded to include inhibitory effects in the graph edges by allowing negative weights or by having separate graphs for excitation and inhibition. Again, as long as the mechanisms shaping these connections do not disrupt a symmetry inherent in the input statistics, the theory remains valid.

Finally, the central claim of this article, namely that invariance transformations are encoded as graph symmetries, is somewhat similar to the proposition that numbers are encoded as binary states of electrical current in a microchip: the representation only has practical value when it is paired with a computational mechanism -- logical gates and circuits in the case of the microchip.
We offer some speculations about how the brain employs graph symmetries in computation in \Cref{sec:discussion:readout}.
But even in complete ignorance of the computational process, a theory of such representations is of value by itself: It can be tested in experiment as shown in \Cref{sec:discussion:predictions} and, if correct, it provides a basis and direction for further investigation.

%---------------------------------------------
\section{Empirical Support} \label{sec:empirical} % 
%-------------------------------------------------------

In the following, we present a survey of empirical observations in support of two major assertions of this paper: Firstly, that the symmetries of concurrence graphs generated by natural sensory perceptions are indeed indicative of real-world invariance transformations. And secondly, that the synaptic connectivity structure in primary sensory cortices approximates the proposed theme of connectivity and thus the concurrence graph.

\subsection{Invariances in Natural Stimuli} \label{sec:empirical:natural}

Empirical support for the idea that invariance transformations of the environment are encoded in the concurrence structure of features is available for both visual and auditory perception.

\paragraph{Vision} A significant number of studies have analyzed the intrinsic statistics of natural images 
(see \eg \cite{jinggang_huang_statistics_1999, ruderman_statistics_1994, geisler_visual_2008} and references therein). A standard approach is to decorrelate the data by decomposing it into features which usually represent small edges of different position and orientation in the image \parencite{olshausen_emergence_1996}. Several studies have computed correlations between such features and consistently found two types of interaction: First, the strongest correlations exist between co-linear features, \ite, between edges which are positioned along a straight line \parencite{August-Zucker_curve_2000, kruger_collinearity_1998, geisler_edge_2001}. Second, a positive correlation is also found between features which are co-circular \parencite{sigman_common_2001}, \ite between edges which are positioned such that one circle can be drawn through both of them, see \Cref{fig:correlations:edges}.
Altogether this shows that features in natural images have an intricate correlation structure depending on their distance and relative orientations, which is a necessary condition  for meaningful symmetries to be identified.

The studies cited above report feature correlations only for relative distances and mostly for relative orientations of features rather than for absolute ones.
For our purpose, this is a limitation as translation and rotation invariance of natural images are then implicitly presumed rather than measured. 

Translation invariance seems to be generally accepted to hold, at least approximately and given that the selection of images is not too narrow\footnote{A set of landscape photos with a blue sky in the upper half will certainly not exhibit translation invariant statistics.}.
It is even more plausible for the statistics of real visual percepts than for collections of photographs, since eye saccades constantly create sequences of translated copies on the retina.

Rotational invariance of image statistics does not hold exactly since natural visual stimuli are somewhat anisotropic \parencite{hansen_horizontal_2004}.
There is a quantitative dominance of horizontal and vertical edges, but according to \parencite{sigman_common_2001} the correlation structure between features is at least qualitatively the same for different (absolute) orientations.
The anisotropy might also by attenuated when taking into account the full visual experience of an animal or human during development, as opposed to a set of photographs taken with a (usually) horizontally aligned camera. 

Scale invariance is another well-studied property of natural images: Several of their statistical properties are not affected by zooming into or out of the picture \parencite{ruderman_origins_1997}.
The set of feature detectors should therefore cover different scales as well, unless a bias for a certain size of receptive fields is inherent in the learning process.
A mix of features with similarly shaped receptive fields of different spatial extension is indeed the outcome of computational models like sparse coding applied to natural images \parencite{olshausen_emergence_1996, olshausen_sparse_1997}.
For real neurons the situation appears to be more complicated: \enquote{the widely accepted notion that receptive fields of neurons in V1 are scaled replica of each other [\dots] is valid in general only to a first approximation} \parencite{teichert_scale-invariance_2007}.
Nevertheless, given that both the image statistics and the set of feature detectors are (at least approximately) scale-invariant, it seems reasonable to assume that the concurrence graph also exhibits the respective symmetry approximately.

In summary, there is strong evidence that the correlation structure between features in natural images is pronounced enough to make the search for symmetries in the concurrence graph meaningful. According to the data presented, it is to be expected that the graph is at least approximately invariant under translation, rotation, or rescaling.

\paragraph{Acoustics} Strong correlations between frequency bands differing by small integer ratios should be expected in a wide variety of sounds, since emitters and resonators tend to mechanically oscillate at a mix of their fundamental frequency and some overtones simultaneously.
Indeed, such correlations have been measured by Abdallah and Plumbley \parencite{Abdallah:2006b} in real music data. \Cref{fig:correlations:frequ} shows a schematic of the cross-correlations between short-term Fourier transform magnitudes from several hours of music radio recording. In another study \parencite{Abdallah:2003}, the same authors first used Independent Component Analysis (ICA) to create a set of basis vectors in an attempt to optimally decorrelate a data set of short music samples. Then they estimated the remaining mutual information between the projections of the audio data onto the different basis vectors. Since most of the basis vectors are well localized in frequency space, each of them can be represented by its center frequency. A plot of the estimated mutual information of pairs of frequencies again resembles \Cref{fig:correlations:frequ}.

\begin{figure}[tb]
    \hfill
        \begin{subfigure}[t]{0.48\textwidth}
        \centering
        \includegraphics[width=2.5in]{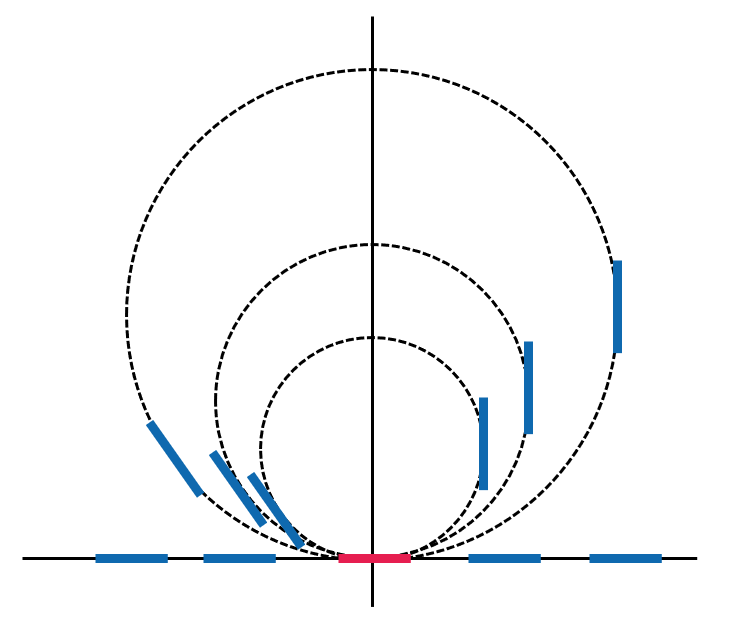}
        \caption{%
            Correlations between edges in natural images: The \ifcolor{blue}{outer} line segments represent the relative positions and orientations of edges with a strong correlation to an edge at the position of the \ifcolor{red}{central} line segment. Correlations are highest for the co-linear edges, followed by the co-circular ones, \cf \parencite{sigman_common_2001}.
        }
        \label{fig:correlations:edges}
    \end{subfigure}
    \hfill
    \begin{subfigure}[t]{0.48\textwidth}
        \centering
        \includegraphics[width=2.5in]{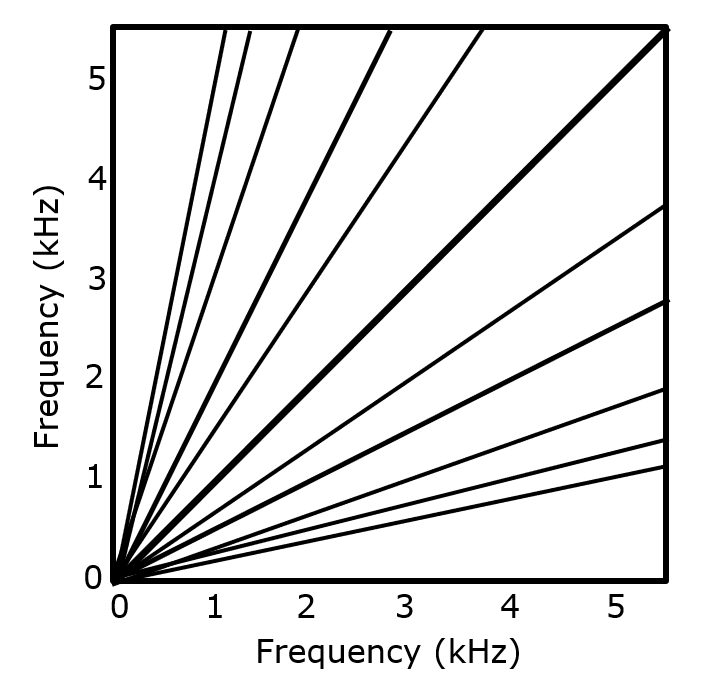}
        \caption{%
            Schematic of the cross-correlations between spectral magnitudes for several hours of music radio. The straight lines represent areas of high correlations. The graph is invariant under a multiplication of the frequency scale. For plots of the original data see \parencite{Abdallah:2003} and \parencite{Abdallah:2006b}.
        }
        \label{fig:correlations:frequ}
    \end{subfigure}
    \caption{Empirical data about feature correlations in natural stimuli.}
\end{figure}

\Cref{fig:correlations:frequ} clearly shows the strong connection between frequencies differing by a harmonic interval.
Considering either the frequency bands or the ICA basis vectors as features (in the sense defined above), the edges of the concurrence graph measure how often two given frequencies contribute significantly and simultaneously to some short audio sample. The plot is therefore qualitatively similar to the weight matrix of the concurrence graph.

An approximate symmetry of the concurrence graph can be found by visual inspection of \Cref{fig:correlations:frequ}: Multiplying every frequency with some constant factor amounts to a rescaling of both axes in each plot, mapping the plot onto a scaled version of itself. Such a rescaling leaves the main features of the plot -- namely the straight lines radiating from the origin -- unchanged, which implies an approximate symmetry of the weight matrix and thus of the concurrence graph. The symmetry is only approximate because it is necessarily broken at very high and very low frequencies. Also, as Abdallah and Plumbley observe, the symmetry is slightly broken by twelve \enquote{ripples} per octave (not shown in \Cref{fig:correlations:frequ}) which seem to be related to the semitone quantization of western music.

In summary, as far as perceptual input is concerned, our theory appears viable both for the visual and the auditory domain. In the following we will show that the concurrence graph -- including its symmetries -- is also possibly implemented in the neural structure of primary sensory cortices.

\subsection{Cortical Theme of Connectivity} \label{sec:empirical:connectivity}
The following is an overview of empirical observations which are consistent with the theme of connectivity described in \Cref{sec:model:concurrencegraph}.

\paragraph{Visual Cortex} Neurons in primary visual cortex are often interpreted as feature detectors.
These cells receive their feed-forward input from the lateral geniculate nucleus and they are activated by features at a particular position and orientation in the visual field.
They are also interconnected through a tight network of recurrent synapses. Several studies \parencite{ko_emergence_2013, iacaruso_synaptic_2017, ko_functional_2011} have shown that two such cells are preferentially connected when their receptive fields are co-oriented and co-axially aligned, thus reflecting the statistical correlation of co-linear edges in natural images. One might expect that the (weaker) correlations between co-circular edges are also expressed in the synaptic connectivity structure, yet the only related study the we are aware of was challenged by very limited data availability and turned out rather inconclusive \parencite{hunt_statistical_2011}. It has been shown though, that the degree of co-circularity in a contour influences human contour detection performance \parencite{geisler_edge_2001}.

\paragraph{Auditory Cortex} Neurons in primary auditory cortex receive feed-forward input from thalamocortical connections as well as intracortical signals via recurrent connections. The feed-forward input is tonotopically organized and A1 neurons typically respond to one or several characteristic frequencies.
According to the hypothetical theme of connectivity, and given the correlation statistics of natural audio stimuli, intra-cortical connections should be strongest between neurons if their characteristic frequencies differ by a harmonic interval. 
Indeed, some support for this hypothesis is reviewed in \parencite{wang_harmonic_2013}:
Tracing the diffusion of a marker substance after local injection into cat auditory cortex shows that \enquote{the intrinsic connections of A1 arising from nearby cylinders of neurons are not homogenous and clusters of cells can be identified by their unique pattern of connections within A1} \parencite{wallace_intrinsic_1991}. In particular, horizontal connections displayed a periodic pattern along the tonotopic
axis. In similar tracing experiments on cat A1 it was found that injections into a specific cortical location caused labeling at other A1 locations that were harmonically related to the injection site \parencite{kadia1999horizontal}.

In summary, evidence from primary sensory cortical areas suggests a common cortical theme of connectivity in which neurons are tuned to specific patterns in their feed-forward input from other brain regions, while being connected intracortically according to statistical correlations between these patterns.

%-------------------------------------------------------
\section{Discussion} \label{sec:discussion}
%-------------------------------------------------------

%-------------------------------------------------------
\subsection{The Read-Out Mechanism} \label{sec:discussion:readout} %
%-------------------------------------------------------

So far we have collected evidence that invariances are encoded in feature correlations of natural stimuli and that they are reproduced in the connectivity structure of primary sensory cortices. An open question remains about the \enquote{read-out mechanism}, \ite how the brain utilizes the concurrence graph to solve computational problems. While a definite answer to this question is out of reach for now, the following arguments indicate that the existence of such a mechanism is indeed conceivable. 

Of course, it is highly improbable that the brain can identify arbitrary graph symmetries without additional assumptions: There is no algorithm known which solves this problem in complete generality in polynomial time, let alone in a biologically plausible way \parencite{kobler_graph_1992}. Yet it is possible that some approximation scheme has evolved which is effective in uncovering those invariances that are encoded in natural stimuli.

Such a heuristic might be based on the assumption that invariance transformations are continuous, \ite they can be generated by a sequence of infinitesimally small steps\footnote{It is not required that such a sequence of infinitesimal transformations can be observed as a time-continuous process.}.
This is certainly true for the important examples of rotation, rescaling and translation in images or multiplicative frequency change in audio signals. Expressed in terms of the concurrence graph, the continuum of transformations is discretized into different permutations of the nodes.
In particular, infinitesimal transformations are approximated by graph automorphisms which map features to only slightly transformed features. Since the feature detectors are not perfectly precise, there will be an overlap of the receptive fields between two detectors whose features differ only by an infinitesimal transformation. 
Such a pair of detectors will be correlated and therefore strongly connected in the concurrence graph.
 
In summary, infinitesimal transformations translate to permutations within neighborhoods of the concurrence graph -- a restriction which dramatically reduces the search space for potential symmetries.

A specific mechanism to exploit these assumptions might rely on wave-like propagation of activity through the network. Suppose that an environmental stimulus activates a certain subset $\Sigma$ of the feature detectors. Assume further that this activation can be passed on to other subsets which are in the graph vicinity of $\Sigma$ and which are the image of $\Sigma$ under a graph automorphism. If each of these subsets can in turn activate further subsets, a wave of activity may propagate along all directions in the space of possible invariance transformations. 
Every point of the wave front contains a transformed representation of $\Sigma$ and as it travels through the network it can be detected by some feature detector on a higher layer.
The wave also maintains the information about the original location of $\Sigma$ in the space of transformations: From the time difference it takes for the wave front to arrive at certain points in transformation space one can always restore the point of origin.
See \Cref{fig:waves} for a sketch of how this mechanism would work in a very simplistic scenario.

One concern about the read-out mechanism sketched above is that the activated subsets of feature detectors might overlap and interfere. To avoid this, their respective activities need to be segregated into different \enquote{channels}. One option is to encode the assignment of detectors to subsets by temporal synchronization of their activity, allowing a single detector to participate in several subsets simultaneously. Another option is to rely on some degree of redundancy in the set of feature detectors, such that each feature can be represented by several detectors which may participate in different subsets independently.

While this account still omits many details, it might lead towards a biologically plausible mechanism to separate a stimulus into a \enquote{what} and a \enquote{where} (either literally or in some abstract space of transformations). 
Wave-like propagation of cortical activity has been observed in many experiments.
In a different article \parencite{powell_can_2021} we proposed that such waves are used to solve planning problems and gave an overview of empirical support for this idea.
It is appealing to speculate that both perception and planning problems which are subject to (invariance) transformations could be supported by essentially the same mechanism.

Finally, using time differences in the sub-millisecond range to infer the location of a stimulus also is known to be in the computational repertoire of the brain, namely when localizing the source of an auditory stimulus \parencite{grothe_mechanisms_2010}. This process is called binaural processing, and it is based on the auditory system's ability to compare the differences in arrival times of sounds at each ear. 

It seems conceivable to employ a similar mechanism to reconstruct the origin of a stimulus in other and more abstract spaces. In fact, time sequence coding has been suggested to be involved in various other brain functions, such as language processing, motor tasks, and vision. In these contexts, the brain must effectively represent and process sequences of information in a timely and organized manner. A comprehensive overview of different models of neural architectures for coding the serial order in sequences is provided by \parencite{Pitti_in_search_2022}. The authors discuss several computational models that attempt to account for the brain's ability to code and represent the temporal structure of sequences, offering insights into the underlying neural mechanisms.

\begin{figure}[tb]
    \centering
    \includegraphics[width=2.5in]{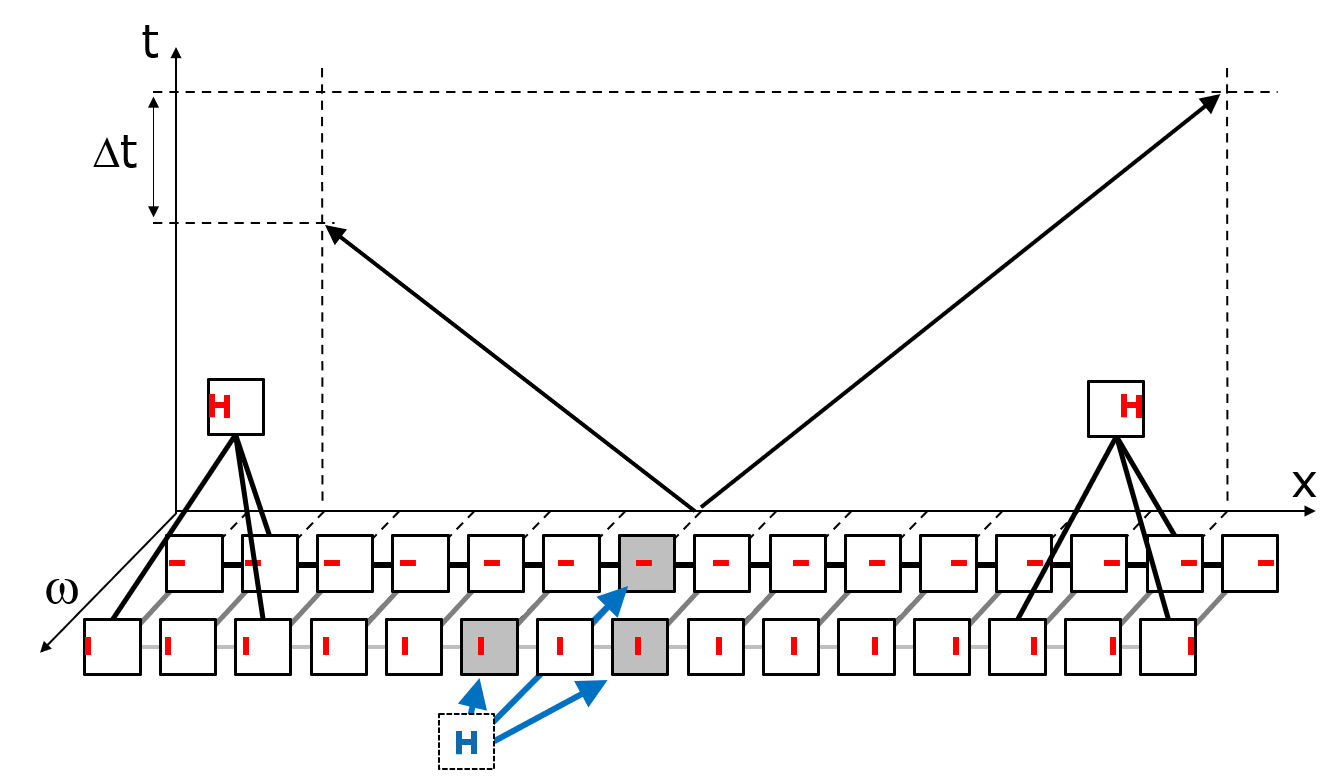}
    \caption{%
        A simple example for a read-out mechanism of the concurrence graph. Feature detectors with two types of orientations $\omega$ have their receptive fields along a \enquote{one-dimensional retina} ($x$-axis) and form a concurrence graph with translational symmetry. When the letter \enquote{H}\ifcolor{(in blue)}{at the bottom}enters the field of vision, it activates three feature detectors (marked in gray). This triggers two wave fronts of activity traveling through the graph, each of which conserves the relative position (as encoded in the graph symmetry) of the three features. The wave fronts reach the two \enquote{H-detectors} on either side and activate them via feed-forward connections. The \enquote{what} information of the stimulus is now encoded in the fact that the H-detectors are activated and the \enquote{where} information is encoded in the time difference $\Delta t$ between their firing (vertical axis, with two black arrows visualizing the spatiotemporal trajectory of the wave fronts). Note that the H-detectors may not have had a chance to emerge as feature detectors on the same layer as the edge detectors in the first place, because the \enquote{H} is less frequent as a pattern than each individual edge. But given the wave propagation dynamics, separate observations of the letter at different positions repeatedly and consistently give rise to the same shape of wave front and thus may be sufficient as a learning signal for new feed-forward feature detectors.
    }
    \label{fig:waves}
\end{figure}

%-------------------------------------------------------
\subsection{Generalized Transformations} \label{sec:discussion:generalization} % 
%-------------------------------------------------------

The concept presented so far is based on global transformations represented by symmetries of the entire concurrence graph. For many perceptual tasks, the brain also requires an understanding of local transformations, such as one individual object moving in front of a static background or several objects moving independently of each other. The potential read-out mechanism outlined above can be extended to allow for such local transformations, too.

In the first step, a global scene activating a feature set is segmented into subsets representing individual objects.
Intriguingly, the concurrence graph itself is well suited as a substrate for a segmentation algorithm: If two features often appear together in general, they are not only likely to have a common cause, but they are also strongly connected in the concurrence graph.
Therefore, when some set of feature detectors are activated simultaneously and their features form a tightly connected clique in the concurrence graph, then they are likely to collectively represent one and the same object.
It has been proposed by Singer that this mechanism of perceptual grouping is implemented in the brain and that the different segments are encoded as independently synchronized assemblies of firing neurons \parencite{singer_synchronization_1993}.
The read-out mechanism proposed in the present article extends Singer's idea by postulating that, in the second step, each of the feature subsets are the source of a wave of activity, traveling through the concurrence graph, and thus encoding for the \enquote{what} and the \enquote{where} of several objects in the scene simultaneously. Effectively, each object locally \enquote{inherits} the transformation which was originally only understood as as a global transformation, induced by a symmetry of the probability distribution $\Psi$, \cf  \Cref{sec:model:problem}.

It remains open if and how the proposed concept can be extended to certain other transformations, with the case of three-dimensional spatial transformations being of particular interest. It is conceivable that progress can be made via a hierarchical stacking of the neural network architecture underlying the present proposal.

%-------------------------------------------------------
\subsection{Other Modalities} \label{sec:discussion:modalities} % 
%-------------------------------------------------------

The preceding discussion was limited to the visual and auditory modalities that provide lucid examples of invariance transformations and for which a large body of empirical results is available.
Yet great care has been taken to avoid any assumptions  specific to those two domains and therefore the concept can in principle be extended to other human or nonhuman modalities.

For tactile sensations the concept of invariance transformations is meaningful, since external objects can be identified through touch regardless of their orientation in space and with some flexibility regarding the body part which makes contact.
Indeed there is evidence that somatosensory cortex follows the theme of connectivity described above with tactile stimuli detectors having receptive fields similar to those in V1 and recurrent connections depending on their likelihood of simultaneous stimulation -- \cf \parencite{powell_can_2021} and references therein.
Yet characterizing the relevant invariances explicitly is harder than in vision or audio due to the interplay between sensory perception and bodily posture.
The author is not aware of any empirical results about the correlation structure of natural tactile stimuli  applicable to the presented concept.

In the olfactory modality \enquote{no obvious metric is available to describe either the space of odor perceptions or the space of odor chemistry} \parencite{wright_odor_2005}.
Consequently, the notion of invariance transformations and thus the presented concept may not apply at all. It fits the picture that olfactory processing also differs anatomically from other modalities in that the respective neural circuits are shallower than their visual and auditory counterparts \parencite{laurent_odor_2001}.

%-------------------------------------------------------
\subsection{Objections} \label{sec:discussion:objections} % 
%-------------------------------------------------------

Several aspects of the presented proposal might be controversial, in particular with respect to the definition of transformations and how they translate into symmetries of the concurrence graph. The following paragraphs address some of the potential concerns.

The basic question how to define invariance transformations invites almost philosophical discussions: Are they characterized by the possibility of some object to undergo the transformation physically like in spatial movement? Or are they rather mappings between hypothetical objects which differ in just one attribute? And if the latter, which attributes are subject to an invariance transformation and which are not? In the present proposal, invariance transformations are simply a way for the brain to replace some na\"ive distance measure in the space of possible perceptions by a metric which is better suited to the respective domain -- suitability being defined by how well it enables the organism to categorize and model its environment, and thus ultimately by evolutionary success. The claim is in essence that symmetries of the concurrence graph allow for statistical learning of transformations whose application in cognitive processes is useful for an animal's survival. There is no need to decide whether some candidate invariance is \enquote{right} or \enquote{wrong}. 

Even our initial assumption of the probability distribution $\Psi$ being symmetric under invariance transformations, see \Cref{sec:model:problem}, is to be understood as a starting point for a simple and consistent mathematical backbone to the proposed concept rather than a definition of real-life invariances. In fact, as critics may point out, $\Psi$ is not invariant under some relevant transformations: Most objects in our environment, for example, have a preferred orientation and thus $\Psi$ cannot be invariant under spatial rotation. Similarly, environmental sound sources and musical instruments alike are restricted to certain frequency ranges and therefore the invariance of $\Psi$ under multiplicative frequency change is broken. Nevertheless, for each object the symmetry may hold approximately within some range (\eg, small angles or frequency multiples close to $1$) and a large number of such objects add to the overall correlation structure of natural stimuli. Given that each of their contributions is transform-invariant over a certain range, it is plausible that the deviations from global transform-invariance approximately balance out and a global symmetry in the concurrence graph emerges. And according to the empirical observations reviewed in \Cref{sec:empirical:natural}, this seems in fact to be the case.

One might still object that a certain symmetry which is manifest in perceptual input could be lost in the observation process. For example, the density of cones and ganglion cells is highly inhomogeneous across the retina \parencite{curcio_topography_1990}, such that even simple spatial translations cannot be expressed as permutations of retina cells. Yet for at least two reasons that observation does not invalidate the presented concept: First, the formation of feature detectors by means of unsupervised learning may restore statistical regularities which are present in the perceptual input data but got distorted during the first stages of processing. For example, the probability of finding an edge at a certain orientation and position in an image (and thus the tendency to develop a detector for that particular edge) should be independent of the resolution at which this image patch is processed locally, as long as it is high enough to clearly represent the edge. Second, if the proposed concept is indeed implemented by biological processes in the brain, then it can be expected to be robust under perturbations. In particular, the read-out mechanism outlined in \Cref{sec:discussion:readout} is based on approximate local symmetries of the graph and might be relatively unaffected by global deviations from perfect symmetry.

Finally, critics may suspect that the projection of $\Psi$ onto lower-dimensional spaces might cause too many artificial symmetries to be useful, \cf \Cref{sec:model:projections}. Yet there is reason to believe otherwise: We have already argued in \Cref{sec:discussion:readout} that neighboring feature detectors have overlapping receptive fields and thus relatively strong mutual connections in the concurrence graph. For simplicity, assume that these neighborhood connections are the strongest ones to be observed at all, which will be true  at least when the density of feature detectors and therefore the overlap between neighbors is high enough. Then neighboring features are always represented by the most strongly connected nodes in the graph and vice versa. Graph symmetries therefore map neighboring features to neighboring features and thus preserve the topology of the feature space, which dramatically reduces the possibilities for symmetries to arise randomly.
Nevertheless, one cannot rule out that artifacts exists and as explained above it is not always obvious whether an invariance is \enquote{real} or an artifact.
For example, one might speculate whether the relative ease and precision with which humans can match musical intervals reflects an evolutionary adaptation facilitating auditory processing or merely a byproduct of the cortical standard mechanism to process topologically arranged stimuli.

%-------------------------------------------------------
\subsection{Comparison to Alternative Concepts} \label{sec:discussion:novelty} % 
%-------------------------------------------------------

Several alternative theories for the emergence of perceptual invariance in the brain have been suggested. Some date back as far as the 1940s but not all of them have passed the test of time, see \parencite{olshausen_neural_2013} for an overview.
It seems to be generally accepted that at least some degree of statistical learning must be involved in establishing invariance transformations, since a complete determination of the relevant neural circuits via evolutionary \enquote{hard-coding} is ruled out by many empirical observations on cortical plasticity \parencite{barnes_sensory_2010}.
The learning mechanism might depend on the particular type of transformation or perceptual modality, but in the light of the anatomical homogeneity and cross-modal plasticity of neocortex one common explanatory framework appears preferable.

Some of the alternative theories are based on the assumption of time continuity and they attempt to reconstruct general transformations from time sequences of perceptions \parencite{foldiak_learning_1991, cadieu_learning_2012}.
These models focus on the the visual system and they do not attempt to explain the emergence of other invariances which can not usually be observed as a time-continuous process, like a change of key in music.

Another class of theories postulates a dynamic remapping of an object's perceptual representation onto some invariant template. Hinton  proposed a neural network with a \enquote{mapping unit} for every possible transformation which sends the present sensory input to its respective  transformed version \parencite{Hinton_Parallel_1981}. The system is designed to optimize the match between the transformed percepts and some previously memorized templates, converging to the right transformation and the right template simultaneously. Hinton's model does not attempt to give a biological explanation for the origin of the mapping units and their correct encoding of invariance transformations.

A different remapping approach based on graph matching has been proposed by \parencite{von_der_malsburg_pattern_statistical_1986} and further developed by von der Malsburg and others \parencite{von_der_malsburg_pattern_1988, lades_distortion_1993, Westphal_feature-driven_2008}. The concept bears some similarity to the ideas presented in this article in that it attempts to identify an object by representing it as a graph of features and matching it to the most similar graph out of a set of memorized templates. In contrast to the present proposal, their graph matching focuses exclusively on the features which are actually observed in a particular image and it ignores their embedding in a wider correlation structure with currently inactive features. The invariance transformations are \enquote{hard-coded} in the graph representation and in the matching process itself, by defining which features at different positions in the image are considered \enquote{the same} and by making the matching explicitly translation invariant or insensitive to rescaling and deformation. Inspired by the graph matching approach, a more recent model \parencite{von_der_malsburg_2015} proposes that the features of an external stimulus and a stored memory are matched via fiber bundles, each of which represents a possible transformation and can be activated or deactivated by a control neuron.

Yet another concept has been put forward by Poggio and Anselmi \parencite{poggio:2016, anselmi_unsupervised_2016}: Given a group $G$ of transformations, a set of template images $t^k$, and all possible transformed templates $gt^k$ ($g \in G$), one can compute a transform-invariant signature for an arbitrary image $I$ with the help of scalar products $\langle I, gt^k\rangle$. For a biological implementation of this idea the authors propose that each $gt^k$ is represented by one \enquote{simple cell} with the appropriate synaptic connections to the pixels of the input image to effectively compute the scalar product. They suggest that this network structure may emerge during visual experience and based on the time continuity assumption (see above). Yet the question remains how a one-dimensional set of temporally consecutive observations can be sufficient to learn the large number of all possible template transformations when the group $G$ is multidimensional, and how robust the image recognition is in situations where $G$ is only partially represented in the template set.  

Finally, processing perceptual data like images, videos or voice is also a very active field of machine learning research and a huge variety of architectures for neural networks have been proposed. Most of them do not focus on biological plausibility but on optimizing performance in practical applications or benchmarks. Learning invariance transformations efficiently and generalizing them to new objects is still a challenge in machine learning. For example, Convolutional Neural Networks have become the most prominent architecture for image recognition, but they are not naturally equivariant to transformations like rescaling or rotating an image \parencite{goodfellow:2016}. This is one important reason why the training of modern deep learning models requires very large amounts of data, as witnessed by the effectiveness of data augmentation techniques where rotated or otherwise transformed images are added to the  training set \parencite{Krizhevsky:2012}. Graph Neural Networks are another class of deep learning models addressing graph-related tasks such as classifying nodes, clustering them or predicting their properties \parencite{wu:2020}. Their main field of application is data which is inherently structured as a graph, like social media profiles with followers or scientific articles citing each other. Computer vision applications of Graph Neural Networks include scene graph generation, point clouds classification, and action recognition.

For convenience, Table~\ref{table:comparison} displays a comparison of the biologically plausible models mentioned in this section.

\begin{table*}[h]
\centering
\caption{Comparison of selected models to explain how the brain might learn invariant representations}
\label{table:comparison}
\begin{tabularx}{\textwidth}{>{\raggedright\arraybackslash}X*{4}{>{\raggedright\arraybackslash}X}}
\toprule
 & Transformation group applied to templates \parencite{poggio:2016} & Learning from natural movies \parencite{cadieu_learning_2012} & Invariant fiber projections \parencite{von_der_malsburg_2015} & Graph symmetries (present proposal) \\
\midrule
Source of knowledge about invariances & Observation of all possible transforms of a \enquote{template} stimulus as a time-continuous process & Observation of objects in motion & By assumption, every homeomorphic mapping of features is an invariance transformation & Concurrence statistics of features in natural stimuli\\
\hline
Learning mechanism & Hebbian learning & Variational learning algorithm & Model for the formation of retinotopic maps & Hebbian learning between feature detectors \\
\hline
Encoding scheme for invariances & Set of templates together with all their possible transforms, each encoded in synaptic weights of \enquote{simple cells} plus one \enquote{complex cell} per template to connect all its transforms & Two-layer neural network: The first represents spatio-temporal features, the second develops form-selective and motion-selective units across multiple features & All possible transformations are stored in the connectivity patterns of control neurons, each of which defines one retinotopic mapping & Graph symmetries in recurrent neural network \\
\hline
Read-out mechanism, i.e., how to apply invariances in problem-solving  & Feed-forward neural network & not discussed & Recurrent convergence process to match stimulus with stored memory & unknown \\
\hline
Modalities & vision & vision & vision & vision, audio, possibly tactile\\
\bottomrule
\end{tabularx}
\end{table*}

% Some kind of wave approach  seems to have been proposed by Lashlay \parencite{lashley1942problem}

%-------------------------------------------------------
\subsection{Predictions} \label{sec:discussion:predictions}
%-------------------------------------------------------

The proposed theory can be tested in experiment: If it is correct, the ability of an organism to apply some invariance transformation to a given perceptual task should depend on its past exposure to stimuli with the corresponding correlation structure.

Investigating the impact of strictly controlled stimuli during developmental stages on the cognitive functions of animals has been a prominent area of research, with its origins dating back to the 1960s. Seminal experiments conducted by Hubel and Wiesel involved occluding the eyes of kittens during development and analyzing the consequences on recordings from the primary visual cortex \parencite{hubel_1970}.

Controlled sensory stimulation has also provided valuable insights into the influence of stimulus statistics on neural development. In one early study, Hirsch and Spinelli reared kittens in such a manner that one eye was solely exposed to horizontal lines, while the other eye was exclusively exposed to vertical lines. Their findings revealed that this manipulation had a profound impact on the development of neurons in the visual cortex. Receptive fields were either horizontally or vertically oriented, and instead of the customary dominance of binocular receptive fields, neurons were predominantly activated by the eye whose past exposure corresponded to the neuron's receptive field orientation \parencite{hirsch_modification_1971}.

Since those early discoveries, numerous researchers have continued to explore the effects of sensory deprivation and manipulation of stimulus statistics on neural plasticity \parencite{espinosa_2012}. The following paragraphs outline similar experiments to test predictions made in the present article.

Assume that two  groups A and B of animals are reared in darkness except for regular visual training cycles during which they are exposed to strictly controlled visual stimuli.
The latter are a set of computer-generated videos which do not contain any time-continuous rotations and which are carefully crafted such that feature correlations are strongly anisotropic (see \Cref{fig:anisotropic} for an example).
While group A is shown those unaltered videos, group B watches every video rotated by a different angle which is chosen at random but constant for the duration of the respective video.
The feature correlations perceived by group B over many videos are thus isotropic.
Lastly, the abilities of all subjects in recognizing rotated visual stimuli are tested.

If rotational invariance were \enquote{hard-coded} in the brain and independent of experience, both groups should complete the final recognition task equally well. 
Yet if invariances were learned by observation of time-continuous transformations, neither group should be able to perform well at the task.
Finally, the concept presented in this article makes the very specific prediction that group B should perform significantly better than group A, because only group B received perceptual input with rotationally invariant correlation structure.
\begin{figure}[tb]
    \centering
    \includegraphics[width=2.5in]{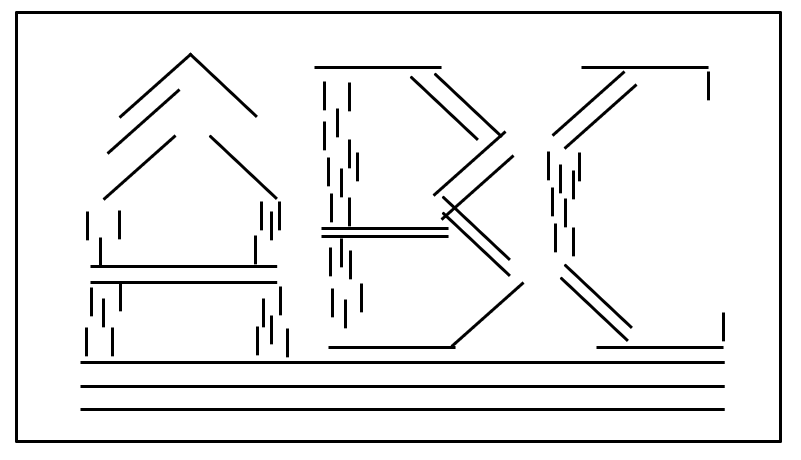}
    \caption{%
        Example of an image with strongly anisotropic correlation structure. Long straight lines, which are the predominant source of strong correlations between collinear features, are preferably oriented in horizontal direction.
    }
    \label{fig:anisotropic}
\end{figure}

Similar experiments can be performed for the auditory modality: The present theory predicts that the ease at which an animal can relate two frequency intervals or melodies depends on its auditory experience during development. Assume that an animal is reared under conditions where all sounds are modified such that the natural correlation structure (\Cref{fig:correlations:frequ}) is replaced by a distorted one. This should have a predictable effect on how quickly the subject can learn equivalence between pairs of auditory stimuli which are connected either by the natural transformation (\ite, multiplication of all frequencies with a constant) or the transformation which corresponds to the distorted correlation structure.

%--------------------------------------------------------
\section{Conclusion} \label{sec:conclusion}
%-------------------------------------------------------

This article established a unified framework describing how the brain might learn a wide range of (not necessarily time-continuous) invariance transformations in multiple sensory modalities without supervision or hardwired domain-specific assumptions.
The proposal explains several seemingly unrelated facts about human perception, \eg the possibility to learn transformations and apply them to new objects or the invariance of musical perception under a change of key, and it makes specific predictions which can be tested in experiments. 
It is consistent with many experimental findings and it is based entirely on basic, biologically plausible mechanisms for the formation of synaptic connectivity.
Depending on the read-out mechanism for the symmetries of the concurrence graph, \cf \Cref{sec:discussion:readout}, the concept may lay the basis for an understanding of abstraction in cognitive processes, \ite the simultaneous classification of a stimulus and its localization in some abstract space.
In order to further solidify the concept, potential read-out mechanisms need to be investigated in more detail. This includes software simulations which may also open the door for new types of brain-inspired artificial intelligence algorithms.

\appendix

%-------------------------------------------------------
\section*{Appendix} \label{appendix}
%-------------------------------------------------------

In this appendix we show that an invariance of the probability distribution ${\Psi: \{0;1\}^n \rightarrow [0;1]}$ gives rise to equivalences between its marginal distributions. We call $\Psi_\mu$ the marginal distribution of the features belonging to some index set $\mu = \{\mu_1,\dots,\mu_m\}$, \ite 
\begin{equation} \label{eq:t1}
    \Psi_\mu(x_{\mu_1},\dots, x_{\mu_m}) = \sum_{\{x_j:\,j \not\in\mu\}} \Psi(x_1, \dots, x_n).
\end{equation}
is the projection of $\Psi$ to the coordinate axes determined by $\mu$. The sum runs over all the $x_j$ which are not selected by the index set $\mu$. The constant $m$ stands for the dimension of the space onto which $\Psi$ is projected with $m=2$ being the most important case for the main text.

We assume that $\Psi$ is invariant under a transformation $T$ , \ite $\Psi(x) = \Psi(Tx)$, and that $T$ can be expressed as a permutation $\tau$ of the coordinate axes, \ite
\begin{equation}\label{eq:tt}
    \Psi(x_1, \dots, x_n) = \Psi(x_{\tau(1)}, \dots, x_{\tau(n)}).
\end{equation}

Then we can show the equivalence between the transformed marginal distributions $\Psi_\mu$ and $\Psi_{\tau(\mu)}$, where $\tau(\mu)$ simply stands for $\{\tau(\mu_1),\dots,\tau(\mu_m)\}$, starting with 
\begin{eqnarray*}  
    &&\Psi_{\tau(\mu)}(x_{\tau(\mu_1)},\dots, x_{\tau(\mu_m)}) \\
    &\stackrel{\text{(\ref{eq:t1})}}{=}& \sum_{\{x_j:\,j \not\in\tau(\mu)\}}  \Psi(x_1, \dots, x_n) \\
     &\stackrel{\text{(\ref{eq:tt})}}{=}& \sum_{\{x_j:\,j \not\in\tau(\mu)\}} \Psi(x_{\tau(1)}, \dots, x_{\tau(n)}).
\end{eqnarray*}

By a re-labeling $x_{\tau(k)} \rightarrow y_k$ of the coordinate axes this can be written as

\begin{equation*}
    \Psi_{\tau(\mu)}(y_{\mu_1},\dots, y_{\mu_m}) = \sum_{\{y_{\tau^{-1}(j)}:\, j \not\in \tau(\mu)\}}  \Psi(y_1, \dots, y_n).
\end{equation*}

Finally, replacing $\tau^{-1}(j)$ by $j'$ we see how the invariance $T$ translates into a equivalence between different marginal distributions:
\begin{eqnarray*}
    \Psi_{\tau(\mu)}(y_{\mu_1},\dots, y_{\mu_m}) &=& \sum_{\{y_{j'}:\, j' \not\in \mu\}} \Psi(y_1, \dots, y_n)\\
    &=&     \Psi_\mu(y_{\mu_1},\dots, y_{\mu_m}).
\end{eqnarray*}

\section*{Acknowledgment}

The author would like to thank Alexander V. Hopp, Robert Klassert, Raul Mure\textcommabelow{s}an, Aleksandar Vu\v{c}kovi\'c, Mathias Winkel and the anonymous reviewers for helpful suggestions which have improved this paper.

\printbibliography

\end{document}